\documentclass[conference]{IEEEtran}
\IEEEoverridecommandlockouts
\usepackage{cite}
\usepackage{amsmath,amssymb,amsfonts}
\usepackage{algorithmic}
\usepackage{graphicx}
\usepackage{textcomp}
\usepackage{xcolor}
\usepackage{amsthm} 
\usepackage{multirow}

\newtheoremstyle{customdef}
  {}                
  {}                
  {\normalfont}     
  {}                
  {\itshape}        
  {:}              
  { }           
  {\thmname{#1}\thmnumber{ #2}\thmnote{ (#3)}}
\theoremstyle{customdef}
\newtheorem{definition}{Definition} 

\usepackage{booktabs} 

\def\BibTeX{{\rm B\kern-.05em{\sc i\kern-.025em b}\kern-.08em
    T\kern-.1667em\lower.7ex\hbox{E}\kern-.125emX}}
\begin{document}

\title{A Biased Nonnegative Block Term Tensor Decomposition Model for Dynamic QoS Prediction
}

\author{\IEEEauthorblockN{Wenjing Liu}
	\IEEEauthorblockA{\textit{College of Computer and Information Science} \\
		\textit{Southwest University}\\
		Chongqing, China \\
		liuww8625@gmail.com}
	\and
	\IEEEauthorblockN{Yujia Lei}
	\IEEEauthorblockA{\textit{School of Finance} \\
		\textit{Southwest University of}\\
		\textit{Political Science and Law}\\
		Chongqing, China \\
		15223361003@163.com}
	\and
	\IEEEauthorblockN{Qu Wang\textsuperscript{*}}
	\IEEEauthorblockA{\textit{College of Computer and Information Science} \\
		\textit{Southwest University}\\
		Chongqing, China \\
		wangquff@gmail.com}
}
\maketitle

\begin{abstract}
With the rapid development of cloud computing and Web services, Quality of Service (QoS) has become a key criterion for service selection and recommendation. Tensor latent feature analysis provides an effective way to model multidimensional QoS data, and most existing QoS prediction methods are mainly based on Canonical Polyadic (CP) decomposition or Tucker decomposition. However, constrained by their inherent structural properties, these methods cannot accurately capture the complex and dynamic dependencies in user-service interactions, which limits their prediction performance. To address this issue, this paper proposes a dynamic QoS prediction framework based on the Biased Nonnegative Block Term Tensor Decomposition Model, termed BNBT. Specifically, the proposed framework is developed from three aspects: (1) block term tensor decomposition is employed to enhance the representation capability of latent feature learning; (2) linear bias terms are incorporated to further improve prediction accuracy; and (3) a tensor-oriented Single Latent Factor-dependent Nonnegative Multiplicative Update (SLF-NMUT) algorithm is designed for efficient parameter estimation. Extensive experiments on real-world QoS datasets demonstrate that the proposed BNBT framework consistently outperforms several state-of-the-art QoS prediction methods in terms of prediction accuracy.
\end{abstract}


\section{Introduction}
The vigorous development of cloud computing \cite{11180920} and Web service technologies has led to a rapid increase in the number of available services, making Quality of Service (QoS) prediction an important problem in cloud service computing \cite{larsson2025hardware}. Since QoS values usually vary across users, services, and time, dynamic QoS data can be naturally organized as a tensor, which provides an effective way to preserve the intrinsic multi-dimensional structure of QoS observations and uncover latent interaction patterns. However, real-world dynamic QoS data usually form a non-standard tensor, which is characterized by high dimensionality and severe incompleteness. As a result, how to accurately predict unknown QoS values from such sparse and irregular historical observations remains a challenging issue \cite{wuAdvancingNonnegativeLatent2022}.

Over the past decade, extensive studies have been conducted on QoS prediction, among which tensor latent feature analysis models have attracted considerable attention due to their ability to capture high-order correlations in multidimensional QoS data \cite{wuPIDincorporatedLatentFactorization2022,liaoNovelTensorCausal2025}. Most existing tensor-based QoS prediction methods are mainly built upon Canonical Polyadic (CP) decomposition \cite{luoTemporalPatternAwareQoS2020} or Tucker decomposition \cite{tangTemporalPatternawareQoS2024,wuLearningAccurateRepresentation2025}. Although these models have achieved promising results, they still suffer from inherent structural limitations. Specifically, CP decomposition represents a tensor as a sum of rank-one components, which is often too restrictive to characterize the complex and dynamic dependencies in user-service interactions. Tucker decomposition is more flexible, but its fully coupled core tensor may introduce excessive model complexity and reduce robustness when dealing with highly sparse non-standard QoS tensors. Consequently, these two decomposition paradigms cannot adequately balance expressive ability and structural compactness, which limits their prediction performance in dynamic QoS scenarios.

To address the above issues, this paper proposes a novel dynamic QoS prediction model, called the Biased Nonnegative Block Term Tensor Decomposition (BNBT) model, which is developed based on block term tensor decomposition with rank-$(L_r \times M_r \times N_r)$. The proposed model is designed to extract more expressive and interpretable latent features from dynamic QoS data so as to improve the prediction accuracy of unknown QoS values. Specifically, BNBT employs block term tensor decomposition to enhance the representation capability of tensor latent feature learning, incorporates linear bias terms to characterize the variation tendencies of QoS data along different dimensions, and develops a tensor-oriented Single Latent Factor-dependent Nonnegative Multiplicative Update (SLF-NMUT) algorithm for efficient parameter estimation. The main contributions of this paper are summarized as follows:

\begin{itemize}
	\item A novel BNBT model is proposed for dynamic QoS prediction, which is developed based on block term tensor decomposition and incorporates linear bias terms to better characterize complex and dynamic user-service interactions;
	\item An efficient optimization algorithm is further developed for BNBT, in which the model parameters are learned through a tensor-oriented nonnegative multiplicative update strategy.
\end{itemize}

\section{Preliminaries}
The user-service-time QoS tensor serves as the fundamental input for dynamic QoS prediction. Therefore, we first formally define the target tensor.

\begin{definition}[\textit{User-Service-Time Tensor for Dynamic QoS}]
	\label{def:hdi}
	Given the user set $I$, service set $J$, and time set $K$ in QoS data, the tensor $\mathbf{Y} \in \mathbb{R}^{|I| \times |J| \times |K|}$ is used to characterize the interaction relationships among users, services, and time. Each element $y_{ijk}$ (where $i \in I, j \in J, k \in K$) denotes the QoS value observed when user $i$ invokes service $j$ at time $k$. Let $\Lambda$ and $\Gamma$ denote the sets of known and unknown entries, respectively. If $|\Lambda| \ll |\Gamma|$, then $\mathbf{Y}$ can be regarded as a non-standard tensor with high dimensionality and incompleteness.
\end{definition}

For a third-order tensor $\mathbf{Y}$, the rank-$(L_r \times M_r \times N_r)$ Block Term Decomposition represents its approximation $\hat{\mathbf{Y}}$ as
\begin{equation}
	\begin{aligned}
		\hat{\mathbf{Y}} = \sum_{r=1}^{R} \mathbf{S}_r \times_1 \mathrm{A}_r \times_2 \mathrm{B}_r \times_3 \mathrm{C}_r,
		\label{eq:btd_approximation}
	\end{aligned}
\end{equation}
where $R$ denotes the number of block terms. $\mathbf{S}_r \in \mathbb{R}^{L_r \times M_r \times N_r}$ is the core tensor of the $r$-th block term. $\mathrm{A}_r \in \mathbb{R}^{|I| \times L_r}$, $\mathrm{B}_r \in \mathbb{R}^{|J| \times M_r}$, and $\mathrm{C}_r \in \mathbb{R}^{|K| \times N_r}$ denote the latent feature matrices of users, services, and time in the $r$-th block term, respectively.

\section{The BNBT Model}
\begin{figure}[tb]
	\centering
	\includegraphics[width=0.4\textwidth]{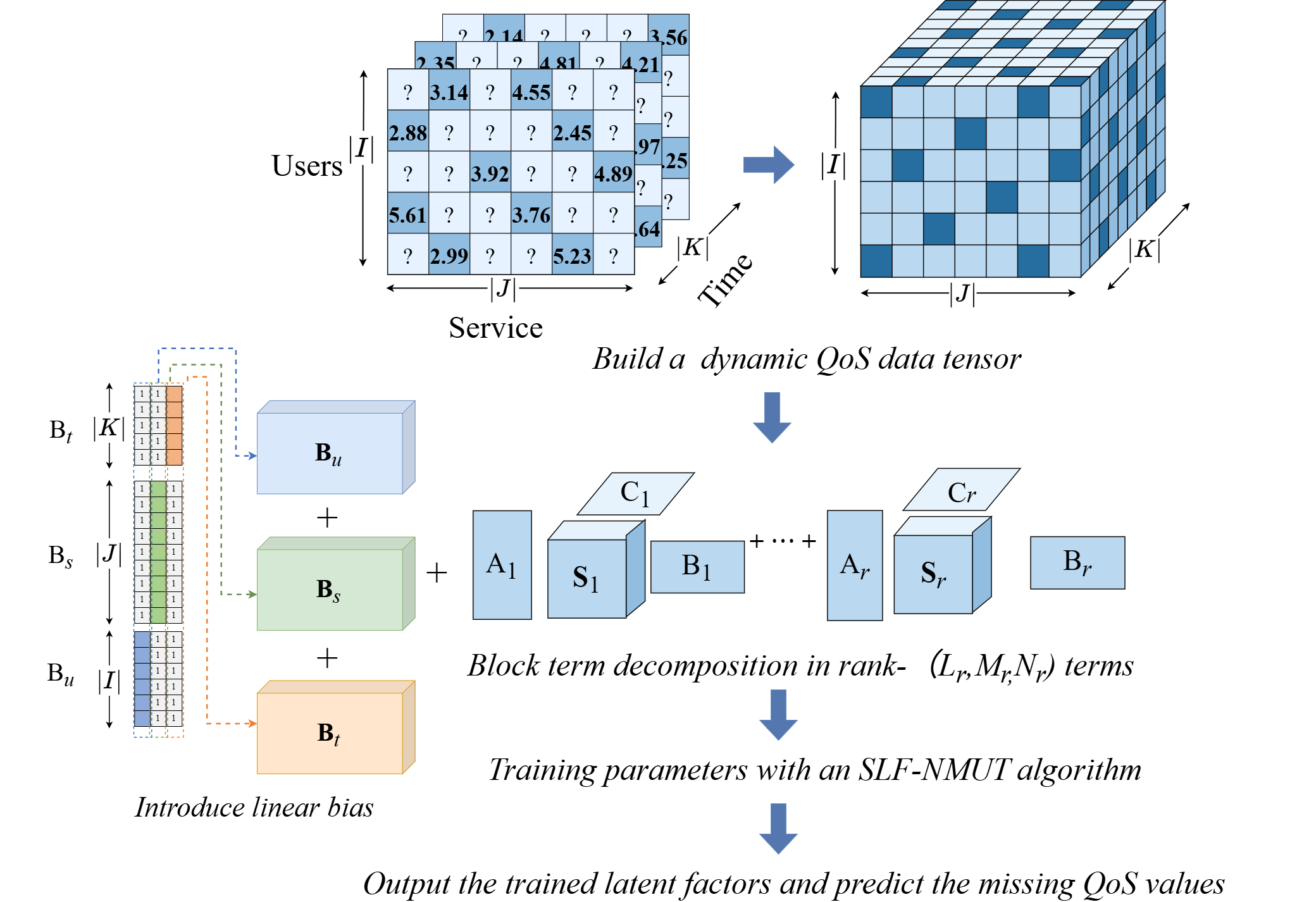}
	\caption{Workflow diagram of the proposed BNBT model.}
	\label{fig:BNBT}
\end{figure}

In this section, we present the methodology of the proposed \emph{Biased Nonnegative Block Term Tensor Decomposition} (BNBT) model. Built upon Block Term Decomposition (BTD), BNBT is designed to capture complex and dynamic user-service interaction patterns from sparse non-standard QoS tensors. In addition, linear bias terms are incorporated to characterize dimension-specific variation tendencies, and a nonnegative multiplicative update strategy is developed for model learning. The overall workflow of BNBT is illustrated in Fig.~\ref{fig:BNBT}.

\subsection{Objective function}
For ease of presentation, let the core tensor set be denoted by $\mathcal{S}=\{\mathbf{S}_1, \mathbf{S}_2, \ldots, \mathbf{S}_R\}$, and let the partitioned factor matrices be defined as $\mathbf{A}=[\mathrm{A}_1, \ldots, \mathrm{A}_R]$, $\mathbf{B}=[\mathrm{B}_1, \ldots, \mathrm{B}_R]$, and $\mathbf{C}=[\mathrm{C}_1, \ldots, \mathrm{C}_R]$. Based on the above definitions and \eqref{eq:btd_approximation}, the element-wise form of the approximation tensor $\hat{\mathbf{Y}}$ can be written as
\begin{equation}
	\begin{aligned}
		\hat{y}_{ijk} = \sum_{r=1}^{R} \sum_{l=1}^{L} \sum_{m=1}^{M} \sum_{n=1}^{N} s_{lmn}^{(r)} a_{il}^{(r)} b_{jm}^{(r)} c_{kn}^{(r)}.
		\label{eq:element_bttd}
	\end{aligned}
\end{equation}

Considering the dynamic nature of QoS data, linear biases can help suppress data fluctuations and further improve prediction accuracy. Therefore, we introduce three linear bias vectors, namely $\boldsymbol{d}$, $\boldsymbol{e}$, and $\boldsymbol{f}$, into the user, service, and time modes of the third-order QoS tensor, respectively \cite{luoTemporalPatternAwareQoS2020}. The element-wise form of the approximation tensor with linear biases is then expressed as
\begin{equation}
	\hat{y}_{ijk} = \sum_{r=1}^{R} \sum_{l=1}^{L} \sum_{m=1}^{M} \sum_{n=1}^{N} s_{lmn}^{(r)} a_{il}^{(r)} b_{jm}^{(r)} c_{kn}^{(r)} + d_i + e_j + f_k.
	\label{eq:element_vector_bias}
\end{equation}

The objective function is constructed by measuring the discrepancy between the target tensor $\mathbf{Y}$ and the approximation tensor $\hat{\mathbf{Y}}$ in terms of the Euclidean distance. Since dynamic QoS data contain a large number of missing entries, only the observed elements are involved in model training. Accordingly, the objective function $\varepsilon$ is defined over the known entry set $\Lambda$. To alleviate overfitting, $L_2$ regularization is incorporated into the objective function. Moreover, since each element in $\mathbf{Y}$ represents a nonnegative QoS value, nonnegativity constraints are imposed on all model parameters. The resulting objective function is defined as follows:
	\begin{equation}
		\small
		\begin{aligned}
			& \varepsilon = \sum_{y_{ijk} \in \Lambda} \Bigg( \Bigg( y_{ijk} - \sum_{r=1}^{R} \sum_{l=1}^{L} \sum_{m=1}^{M} \sum_{n=1}^{N} s_{lmn}^{(r)} a_{il}^{(r)} b_{jm}^{(r)} c_{kn}^{(r)} \\
			& \qquad - d_i - e_j - f_k \Bigg)^2 + \lambda_1 \sum_{r=1}^{R} \sum_{l=1}^{L} \sum_{m=1}^{M} \sum_{n=1}^{N} (s_{lmn}^{(r)})^2 \\
			& \qquad + \lambda_2 \left( \sum_{r=1}^{R} \sum_{l=1}^{L} (a_{il}^{(r)})^2 + \sum_{r=1}^{R} \sum_{m=1}^{M} (b_{jm}^{(r)})^2 + \sum_{r=1}^{R} \sum_{n=1}^{N} (c_{kn}^{(r)})^2 \right) \\
			& \qquad + \lambda_3 (d_i^2 + e_j^2 + f_k^2) \Bigg) \\
			& \text{s.t. } \forall i \in I, j \in J, k \in K, l \in \{1,2,\ldots,L\}, m \in \{1,2,\ldots,M\}, \\
			& \qquad n \in \{1,2,\ldots,N\}, r \in \{1,2,\ldots,R\}: \\
			& s_{lmn}^{(r)} \ge 0,\; a_{il}^{(r)} \ge 0,\; b_{jm}^{(r)} \ge 0,\; c_{kn}^{(r)} \ge 0,\; d_i \ge 0,\; e_j \ge 0,\; f_k \ge 0,
		\end{aligned}
		\label{eq:objective_function_fal}
	\end{equation}
where $\lambda_1$, $\lambda_2$, and $\lambda_3$ denote the regularization coefficients.

\subsection{Optimization}
Since the objective function in \eqref{eq:objective_function_fal} is subject to nonnegativity constraints, it is necessary to ensure that all parameters remain nonnegative during the optimization process. The Single Latent Factor-dependent Nonnegative Multiplicative Update for Tensors (SLF-NMUT) algorithm has been shown to be an efficient optimization strategy for latent factor tensor models \cite{luoTemporalPatternAwareQoS2020}. Motivated by this idea, we develop tensor-oriented single latent factor-dependent nonnegative multiplicative update rules for BNBT. 

First, according to the principle of SLF-NMUT, the partial derivatives of the latent factors in the objective function are derived to obtain the following additive update rules. Since the update rules for $\mathbf{A}$, $\mathbf{B}$, and $\mathbf{C}$ are analogous, only $\mathbf{A}$ is taken as a representative for illustration. Similarly, among the three bias vectors $\boldsymbol{d}$, $\boldsymbol{e}$, and $\boldsymbol{f}$, only $\boldsymbol{d}$ is presented as a representative case:
\begin{equation}
	\small
	\begin{aligned}
		& s_{lmn}^{(r)} \leftarrow s_{lmn}^{(r)} - \eta_{lmn} \sum_{y_{ijk} \in \Lambda} \left( \lambda_1 s_{lmn}^{(r)} - \delta a_{il}^{(r)} b_{jm}^{(r)} c_{kn}^{(r)} \right), \\
		& a_{il}^{(r)} \leftarrow a_{il}^{(r)} - \eta_{il} \sum_{y_{ijk} \in \Lambda(i)} \left( \lambda_2 a_{il}^{(r)} - \delta \sum_{m=1}^{M} \sum_{n=1}^{N} s_{lmn}^{(r)} b_{jm}^{(r)} c_{kn}^{(r)} \right), \\
		& d_i \leftarrow d_i - \eta_i \sum_{y_{ijk} \in \Lambda(i)} \left( \lambda_3 d_i - \delta \right),
	\end{aligned}
	\label{eq:update_rules}
\end{equation}
where $\delta = y_{ijk} - \hat{y}_{ijk}$ denotes the prediction error, $\eta$ denotes the learning rate of the corresponding latent factor, and $\Lambda(i)$ is the subset of $\Lambda$ consisting of the observed entries associated with user $i$. According to the principle of SLF-NMUT\cite{luoTemporalPatternAwareQoS2020}, the negative terms in \eqref{eq:update_rules} can be eliminated by choosing suitable learning rates. Specifically, the learning rates are set as
\begin{equation}
	\small
	\begin{alignedat}{2}
		&\eta_{lmn} &&= s_{lmn}^{(r)} \Bigg/ \sum_{y_{ijk} \in \Lambda} \left( \lambda_1 s_{lmn}^{(r)} + \hat{y}_{ijk} a_{il}^{(r)} b_{jm}^{(r)} c_{kn}^{(r)} \right), \\
		&\eta_{il} &&= a_{il}^{(r)} \Bigg/ \sum_{y_{ijk} \in \Lambda(i)} \left( \lambda_2 a_{il}^{(r)} + \hat{y}_{ijk} \sum_{m=1}^{M} \sum_{n=1}^{N} s_{lmn}^{(r)} b_{jm}^{(r)} c_{kn}^{(r)} \right), \\
		&\eta_i &&= d_i \Bigg/ \sum_{y_{ijk} \in \Lambda(i)} \left( \lambda_3 d_i + \hat{y}_{ijk} \right).
	\end{alignedat}
	\label{eq:learning_rates_group}
\end{equation}

Based on the above derivation, the final multiplicative update rules can be obtained as
\begin{equation}
	\small
	\begin{aligned}
		s_{lmn}^{(r)} &\leftarrow s_{lmn}^{(r)} \frac{\sum\limits_{y_{ijk} \in \Lambda} y_{ijk} a_{il}^{(r)} b_{jm}^{(r)} c_{kn}^{(r)}}{\sum\limits_{y_{ijk} \in \Lambda} \hat{y}_{ijk} a_{il}^{(r)} b_{jm}^{(r)} c_{kn}^{(r)} + \lambda_1 |\Lambda| s_{lmn}^{(r)}}, \\
		a_{il}^{(r)} &\leftarrow a_{il}^{(r)} \frac{\sum\limits_{y_{ijk} \in \Lambda(i)} y_{ijk} \sum\limits_{m=1}^{M} \sum\limits_{n=1}^{N} s_{lmn}^{(r)} b_{jm}^{(r)} c_{kn}^{(r)}}{\sum\limits_{y_{ijk} \in \Lambda(i)} \hat{y}_{ijk} \sum\limits_{m=1}^{M} \sum\limits_{n=1}^{N} s_{lmn}^{(r)} b_{jm}^{(r)} c_{kn}^{(r)} + \lambda_2 |\Lambda(i)| a_{il}^{(r)}}, \\
		d_i &\leftarrow d_i \frac{\sum\limits_{y_{ijk} \in \Lambda(i)} y_{ijk}}{\sum\limits_{y_{ijk} \in \Lambda(i)} \hat{y}_{ijk} + \lambda_3 |\Lambda(i)| d_i}.
	\end{aligned}
	\label{eq:multiplicative_update}
\end{equation}

\subsection{Algorithm design and analysis}

\begin{table}[tb]
	\centering
	\resizebox{\linewidth}{!}{
		\begin{tabular}{ll}
			\toprule
			\multicolumn{2}{l}{\textbf{Algorithm 1} BNBT} \\
			\midrule
			\multicolumn{2}{l}{\textbf{Input:} $\Lambda, I, J, K, R, L, M, N, \lambda_1, \lambda_2, \lambda_3$} \\
			\midrule
			\textbf{Operation} & \textbf{Cost} \\
			\midrule
			1. Initialize $\mathcal{S}, \mathbf{A}, \mathbf{B}, \mathbf{C}, \boldsymbol{d}, \boldsymbol{e}, \boldsymbol{f}$ randomly in $[0, 0.05]$ & \begin{tabular}{@{}l@{}}$\Theta(|I|LR+|J|MR+|K|NR$ \\ $+LMNR+|I|+|J|+|K|)$\end{tabular} \\
			2. Initialize $n=1, T=max\_iter$ & $\Theta(1)$ \\
			\textbf{while} $n \le T$ \textbf{and} not converge \textbf{do} & $\times n$ \\
			3. Update $\mathcal{S}$ with ~\eqref{eq:multiplicative_update} & $\Theta(|\Lambda|RLMN + RLMN)$ \\
			4. Update $\mathbf{A}$ with ~\eqref{eq:multiplicative_update} & $\Theta(|\Lambda|RLMN + |I|RL)$ \\
			5. Update $\mathbf{B}$ with ~\eqref{eq:multiplicative_update} & $\Theta(|\Lambda|RLMN + |J|RM)$ \\
			6. Update $\mathbf{C}$ with ~\eqref{eq:multiplicative_update} & $\Theta(|\Lambda|RLMN + |K|RN)$ \\
			7. Update $\boldsymbol{d}$ with ~\eqref{eq:multiplicative_update} & $\Theta(|\Lambda| + |I|)$ \\
			8. Update $\boldsymbol{e}$ with ~\eqref{eq:multiplicative_update} & $\Theta(|\Lambda| + |J|)$ \\
			9. Update $\boldsymbol{f}$ with ~\eqref{eq:multiplicative_update} & $\Theta(|\Lambda| + |K|)$ \\
			\textbf{end while} & \\
			\midrule
			\multicolumn{2}{l}{\textbf{Output:} $\mathcal{S}, \mathbf{A}, \mathbf{B}, \mathbf{C}, \boldsymbol{d}, \boldsymbol{e}, \boldsymbol{f}$} \\
			\bottomrule
	\end{tabular}}
\end{table}

Based on the above derivations, the proposed BNBT algorithm is summarized in Algorithm 1, whose objective is to learn the latent factors $\mathcal{S}$, $\mathbf{A}$, $\mathbf{B}$, $\mathbf{C}$, $\boldsymbol{d}$, $\boldsymbol{e}$, and $\boldsymbol{f}$ for missing QoS prediction. The algorithm consists of seven sub-procedures, namely $\mathbf{Update\_S}$, $\mathbf{Update\_A}$, $\mathbf{Update\_B}$, $\mathbf{Update\_C}$, $\mathbf{Update\_d}$, $\mathbf{Update\_e}$, and $\mathbf{Update\_f}$. For simplicity, $L$, $M$, and $N$ are uniformly used to denote the dimensions of the local latent feature spaces for all block terms. According to Algorithm 1, the time and space complexities of BNBT are approximately $\Theta(n|\Lambda|RLMN)$ and $\Theta(|\Lambda| + R \times (|I|L + |J|M + |K|N + LMN))$, respectively. Therefore, BNBT has linear complexity with respect to the number of observed entries and exhibits favorable computational efficiency and scalability.

\section{Experiments}

\subsection{Experimental Settings}

Experiments are conducted on two publicly available real-world dynamic QoS datasets \cite{zhengInvestigatingQoSRealWorld2014a}, namely D1 for response time and D2 for throughput. Each dataset contains QoS observations of 142 users on 4,500 services over 64 continuous time slices. Both datasets have 30,287,611 service invocation records and the same densities of 74.06\%. To evaluate the prediction performance under different sparsity levels, each dataset is further divided into four sub-datasets with training, validation, and test ratios of 10\%:10\%:80\%, 20\%:10\%:70\%, 30\%:10\%:60\%, and 60\%:10\%:30\%, respectively, denoted by D1.1--D1.4 and D2.1--D2.4. The detailed dataset statistics are summarized in Table~\ref{tab:data_setting}, where $\Omega$, $\Psi$, and $\Phi$ denote the training, validation, and test sets, respectively.

\begin{table}[t]
	\centering
	\caption{Statistics of the dynamic QoS datasets.}
	\label{tab:data_setting}
	\begin{tabular}{lcccc}
		\toprule
		Dataset & QoS type & Users & Services & Time slices  \\
		\midrule
		D1 & Response time & 142 & 4,500 & 64 \\
		D2 & Throughput    & 142 & 4,500 & 64 \\
		\bottomrule
	\end{tabular}
\end{table}

The prediction performance is evaluated by Root Mean Square Error (RMSE) and Mean Absolute Error (MAE), defined as
{\small
	\begin{equation}
		\begin{aligned}
			\mathrm{RMSE} &= \sqrt{\frac{1}{|\Phi|} \sum_{y_{ijk} \in \Phi} (y_{ijk} - \hat{y}_{ijk})^2}, \\
			\mathrm{MAE}  &= \frac{1}{|\Phi|} \sum_{y_{ijk} \in \Phi} |y_{ijk} - \hat{y}_{ijk}|.
		\end{aligned}
	\end{equation}
}

\subsection{Comparison with other models}

To verify the effectiveness of BNBT, we compare it with the following representative dynamic QoS prediction models:

\begin{itemize}
	\item \textbf{M1}: A CP-based latent feature tensor model with biased learning objective and SLF-NMUT optimization \cite{luoTemporalPatternAwareQoS2020}.
	\item \textbf{M2}: A biased nonnegative Tucker tensor decomposition model \cite{tangTemporalPatternawareQoS2024}.
	\item \textbf{M3}: The proposed BNBT model based on Block Term Tensor Decomposition.
\end{itemize}

For a fair comparison, all methods are evaluated under the same experimental protocol. The latent feature dimension is set to $R=3$ for all compared models, while the remaining hyperparameters are tuned individually for their best performance. The maximum number of training iterations is set to 1,000, and the algorithm is regarded as converged when the error change between two consecutive iterations is smaller than $10^{-5}$. To reduce the influence of random initialization, all models are initialized under the same setting and are independently run 10 times.

\begin{figure}[htbp]
	\centering
	\includegraphics[width=0.48\linewidth]{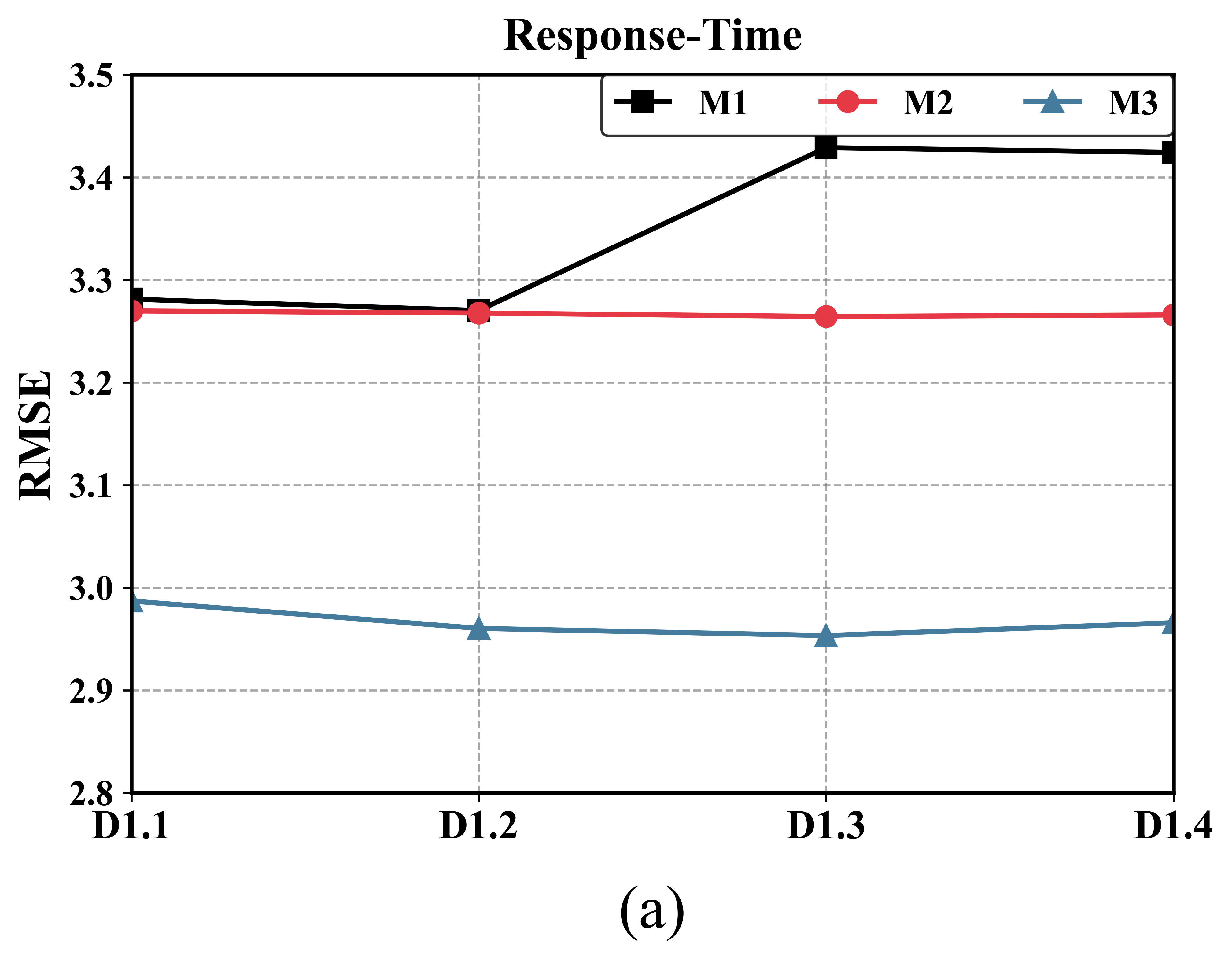} \hfill
	\includegraphics[width=0.48\linewidth]{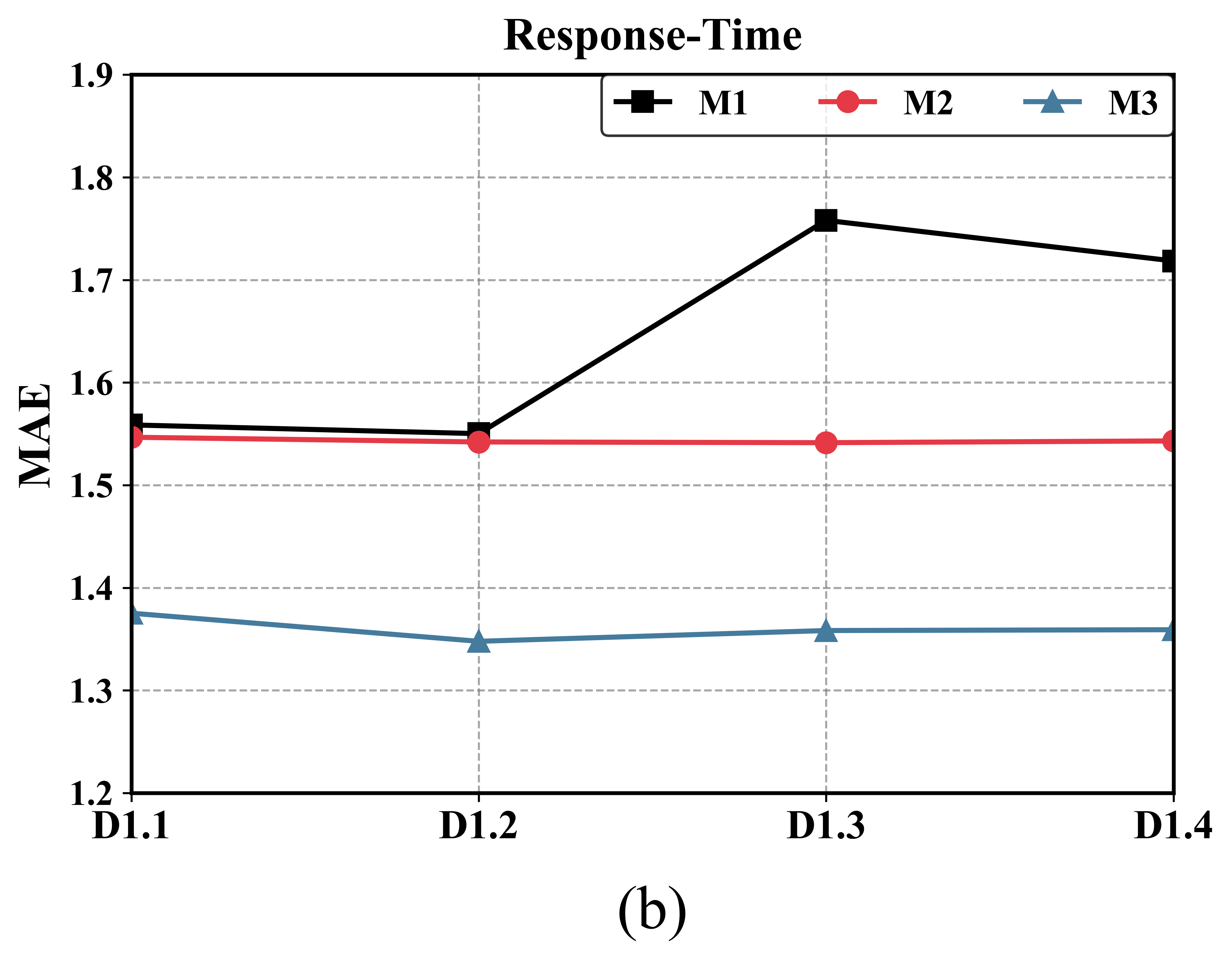}
	\includegraphics[width=0.48\linewidth]{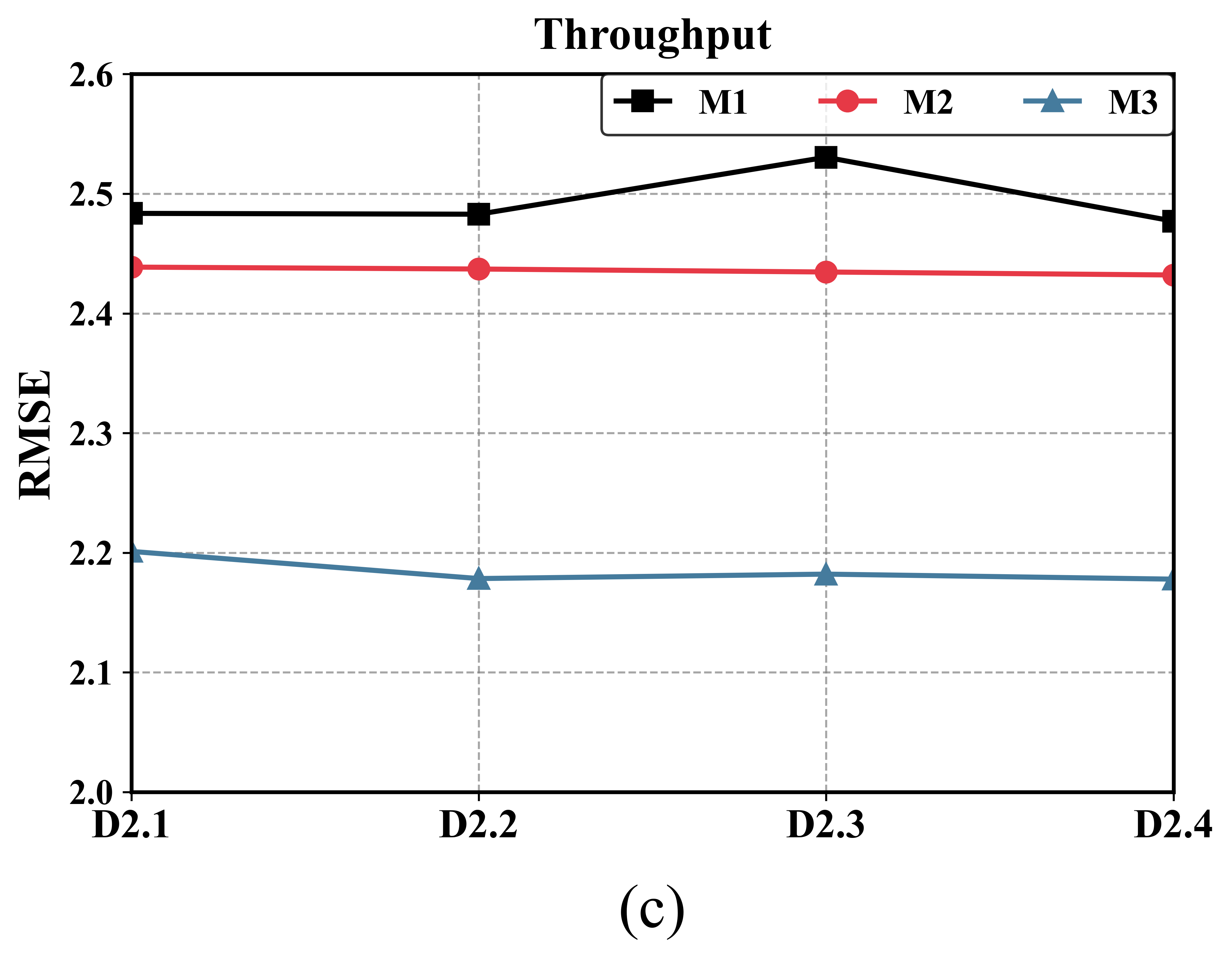} \hfill
	\includegraphics[width=0.48\linewidth]{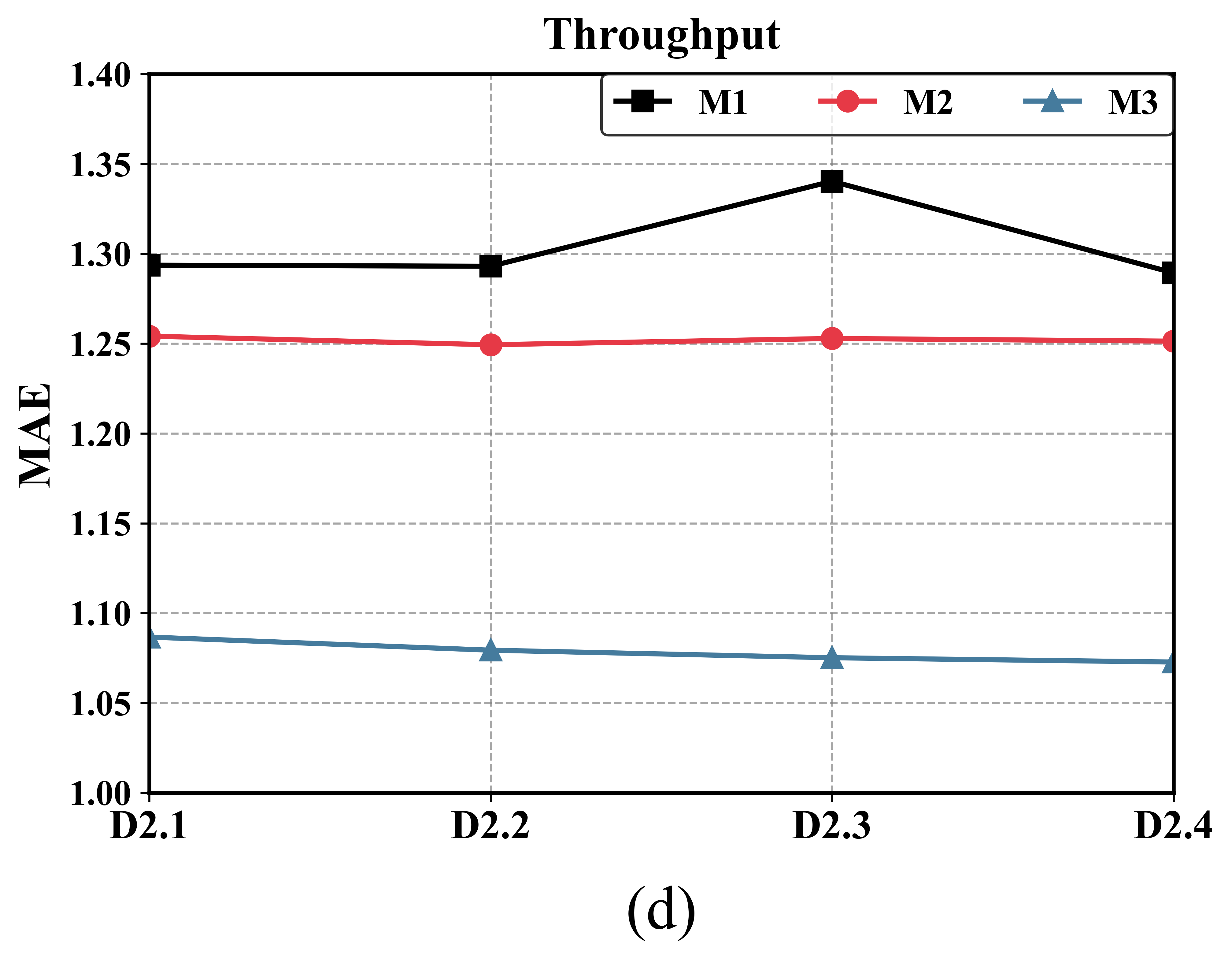}
	\caption{Prediction performance of different models under varying data densities.}
	\label{fig:results}
\end{figure}

The comparison results are presented in Fig.~\ref{fig:results}, from which several observations can be drawn. First, BNBT consistently achieves the lowest RMSE and MAE across all experimental cases on both D1 and D2, demonstrating its superior performance for dynamic QoS prediction. In particular, its advantage is more pronounced under relatively sparse settings such as D1.1 and D2.1, indicating that BNBT has stronger robustness in handling sparse non-standard QoS tensors. Second, BNBT consistently maintains the best performance under varying training data densities, showing that its superiority mainly comes from its more expressive model structure rather than merely from denser observations. Therefore, BNBT is a promising model for QoS prediction or other complex data-related industrial applications. Compared with the CP-based M1 and the Tucker-based M2, BNBT better balances representation capability and structural compactness through Block Term Decomposition, thereby capturing more complex and dynamic user-service interaction patterns. For example, on D1.1, the RMSE of BNBT is 2.9871, which is 8.97\% and 8.65\% lower than those of M1 and M2, respectively, while its MAE is 1.3752, corresponding to improvements of 11.77\% and 11.09\%, respectively.

\section{Conclusion}

To more effectively capture the complex and dynamic interaction patterns hidden in QoS data, this paper proposes a dynamic QoS prediction model, called BNBT. Compared with existing tensor latent feature-based QoS prediction frameworks, the proposed BNBT model is developed based on rank-$(L_r \times M_r \times N_r)$ Block Term Decomposition, which provides stronger representation capability for modeling dynamic QoS tensors. In addition, linear bias terms are incorporated to further improve prediction accuracy, and the proposed framework is optimized via the SLF-NMUT algorithm to ensure nonnegativity during the training process. Experimental results on two publicly available QoS datasets demonstrate that BNBT consistently outperforms several representative baseline methods. Although BNBT has shown promising performance, there is still room for further validation and improvement. In future work, a comprehensive ablation study will first be conducted to specifically isolate and quantify the impact of the linear bias terms on the overall prediction accuracy. Furthermore, adaptive hyperparameter tuning techniques will be explored to avoid laborious manual adjustments, and the training efficiency of the proposed model can be further enhanced by introducing stochastic gradient-based optimization strategies. 

\bibliographystyle{IEEEtran}
\bibliography{Reference}

\end{document}